\documentclass[10pt,twocolumn,letterpaper]{article}

\usepackage[pagenumbers]{cvpr} %

\usepackage{amsmath}
\usepackage{animate}
\usepackage{subcaption}
\usepackage{array}
\usepackage{multirow}
\usepackage{microtype}
\usepackage{enumitem}
\usepackage{xspace}
\usepackage{xcolor}
\usepackage{soul}
\usepackage{multicol}
\usepackage{colortbl}
\usepackage{tabularx,booktabs}
\usepackage{cancel}
\usepackage{booktabs}
\usepackage{float}
\usepackage{pifont}
\usepackage{adjustbox}
\usepackage{epigraph}
\usepackage{wrapfig}
\usepackage{textcomp}  %
\usepackage{scalerel}  %
\usepackage{duckuments}
\usepackage{makecell}

\newcommand{\sota}{state-of-the-art\xspace}

\def\rocket{\scalerel*{\scalebox{-1}[1]{\includegraphics{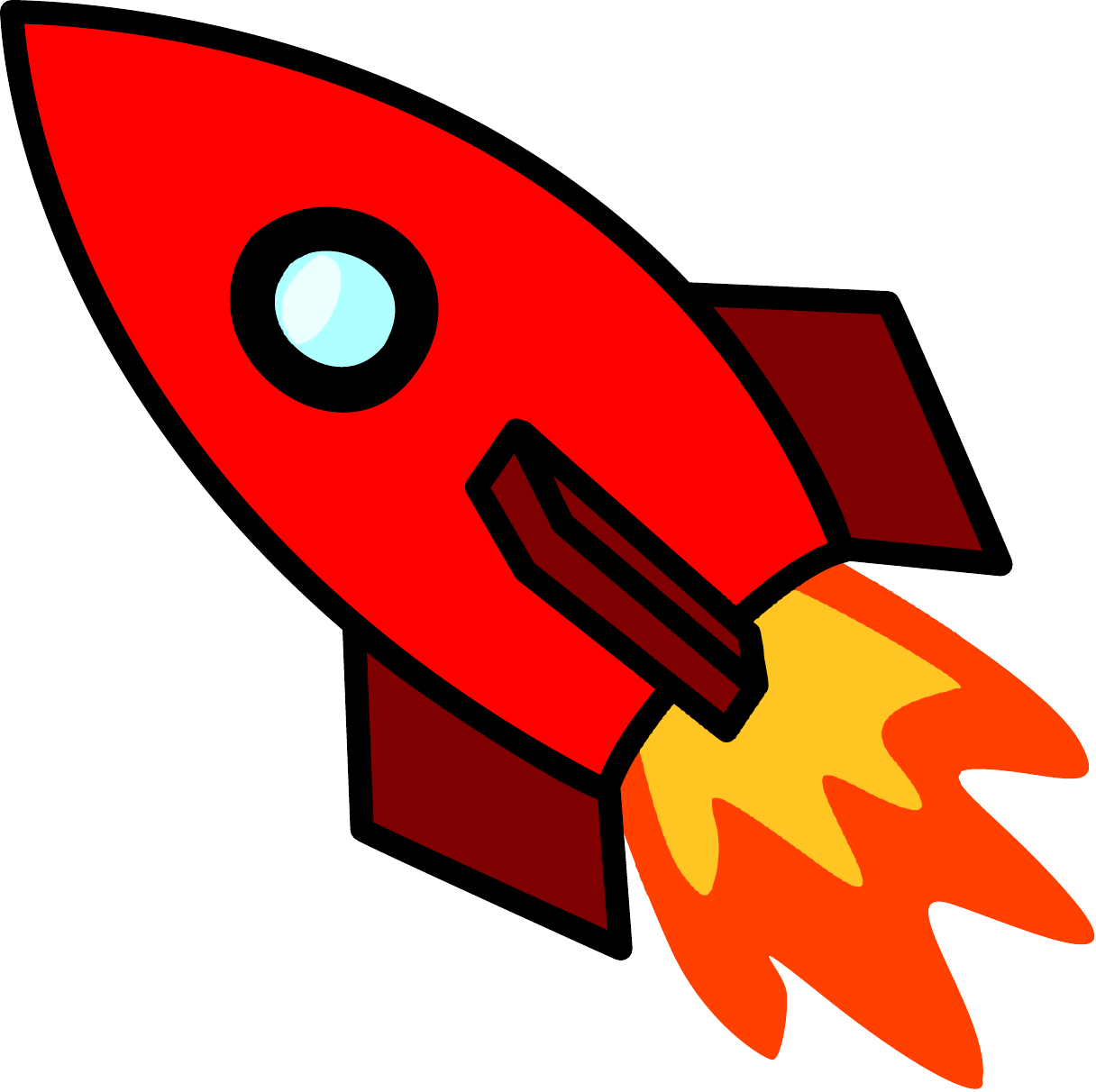}}}{\Huge\textrm{\textbigcircle}}}

\makeatletter
\DeclareRobustCommand\onedot{\futurelet\@let@token\@onedot}
\def\@onedot{\ifx\@let@token.\else.\null\fi\xspace}

\def\etc{\emph{etc}\onedot} 
 
\def\etal{\emph{et al}\onedot}
\makeatother

\let\oldFootnote\footnote
\newcommand\nextToken\relax
\renewcommand\footnote[1]{%
    \oldFootnote{#1}\futurelet\nextToken\isFootnote}
\newcommand\isFootnote{%
    \ifx\footnote\nextToken\textsuperscript{,}\fi}

\definecolor{overview_orange}{HTML}{FFA726}
\definecolor{overview_green}{HTML}{9CCC65}
\definecolor{overview_purple}{HTML}{D877E9}
\colorlet{overview_red}{red!35}
\definecolor{overview_gray}{HTML}{BEBEBE}

\newcommand{\acrobat}{\emph{Click any result to play video.}}

\newcolumntype{Y}{>{\centering\arraybackslash}X}

\newcommand{\PreserveBackslash}[1]{\let\temp=\\#1\let\\=\temp}
\newcolumntype{C}[1]{>{\PreserveBackslash\centering}p{#1}}
\newcolumntype{R}[1]{>{\PreserveBackslash\raggedleft}p{#1}}
\newcolumntype{L}[1]{>{\PreserveBackslash\raggedright}p{#1}}

\newcolumntype{F}[1]{%
 >{\vbox to 5ex\bgroup\vfill\centering}%
 p{#1}%
 <{\egroup}}

\usepackage{pifont}
\newcommand{\cmark}{\ding{51}}
\newcommand{\xmark}{\ding{55}}
\definecolor{Green}{RGB}{13, 212, 66}

\definecolor{Gray}{gray}{0.92}
\newcolumntype{a}{>{\columncolor{Gray}}c}

\makeatletter
\let\UrlSpecialsOld\UrlSpecials
\def\UrlSpecials{\UrlSpecialsOld\do\/{\Url@slash}\do\_{\Url@underscore}}%
\def\Url@slash{\@ifnextchar/{\kern-.11em\mathchar47\kern-.2em}%
    {\kern-.0em\mathchar47\kern-.08em\penalty\UrlBigBreakPenalty}}
\def\Url@underscore{\nfss@text{\leavevmode \kern.06em\vbox{\hrule\@width.3em}}}
\makeatother
\newcommand{\ours}{SPACE\xspace}

\newcommand{\speechtolandmarks}{Speech2Landmarks\xspace}
\newcommand{\landmarkstolatents}{Landmarks2Latents\xspace}

\usepackage{graphicx}
\usepackage{amsmath}
\usepackage{amssymb}
\usepackage{booktabs}

\usepackage[pagebackref,breaklinks,colorlinks]{hyperref}

\usepackage[capitalize]{cleveref}
\crefname{section}{Sec.}{Secs.}
\Crefname{section}{Section}{Sections}
\Crefname{table}{Table}{Tables}
\crefname{table}{Tab.}{Tabs.}

\begin{document}

\title{SPACE \rocket: Speech-driven Portrait Animation with Controllable Expression}

\author{Siddharth Gururani, Arun Mallya, Ting-Chun Wang, Rafael Valle, Ming-Yu Liu\\[3pt]
NVIDIA\\
{\tt\small \{sgururani, amallya, tingchunw, rafaelvalle, mingyul\}@nvidia.com}
}

\twocolumn[{%
\renewcommand\twocolumn[1][]{#1}%
\maketitle
\begin{center}
    
\vspace{-20pt}
\centering

\adjustbox{max width=0.95\textwidth,center}{
\setlength{\tabcolsep}{0pt}
\centering
\hspace{-8pt}
\begin{tabular}{cC{0.25\textwidth}C{0.25\textwidth}C{0.25\textwidth}C{0.25\textwidth}}
    & \multicolumn{4}{c}{\href{https://deepimagination.cc/SPACE/data/teaser/image_id03862-JPpFLTSa6iU-00182_audio_id04295-7_t4qaydG3Y-00026.mp4}{\includegraphics[width=\textwidth]{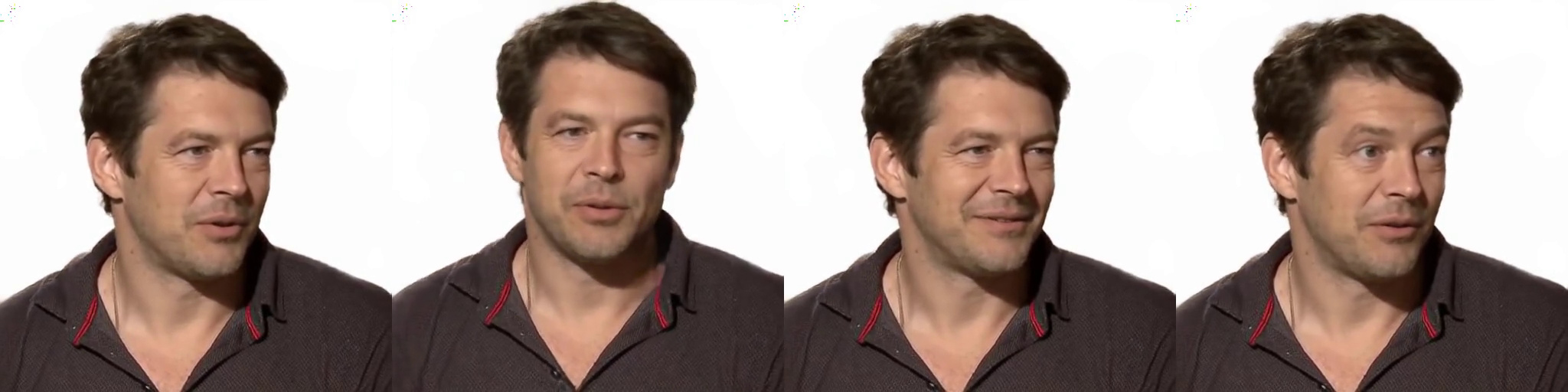}}} \\[-4pt]
    & \multicolumn{4}{c}{\href{https://deepimagination.cc/SPACE/data/teaser/image_woman_audio_id04570-9dAsIrZc5RY-00077.mp4}{\includegraphics[width=\textwidth]{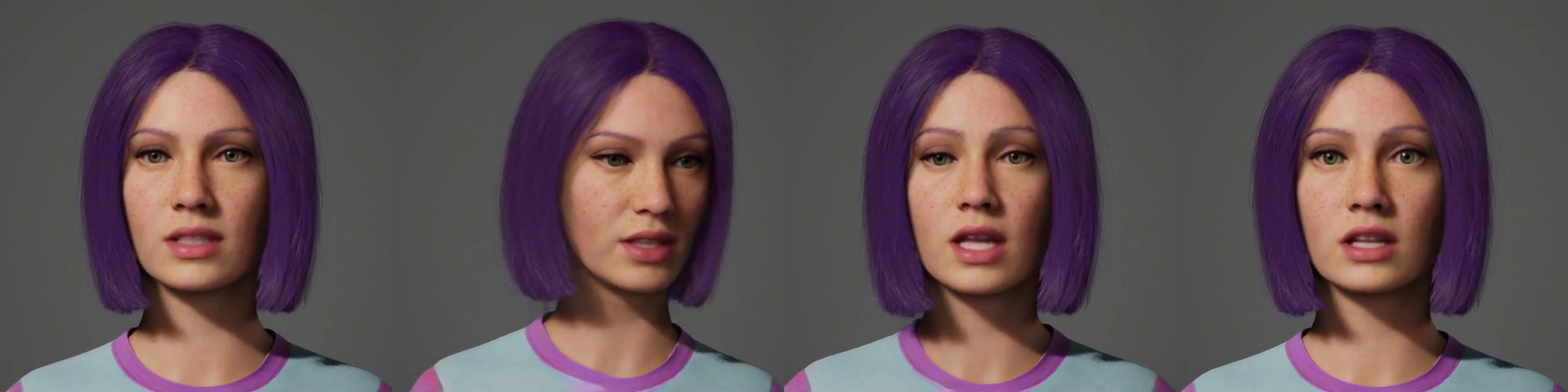}}} \\[-4pt]
    \midrule
    \textbf{Pose} & Generated & Transferred & Generated & Generated \\
    \midrule
    \textbf{Emotion} & Neutral & Neutral & Happy (top) / Sad (bottom) & Surprise (top) / Fear (bottom) \\
    \midrule
\end{tabular}}
\vspace{-10pt}
\captionof{figure}{\textbf{\ours} enables speech-driven animation of a portrait, with control over the output pose, emotions, and intensities of expressions. It can handle a wide range of input poses and produce realistic and high-resolution outputs. \acrobat
}
\label{fig:teaser}

\end{center}%
}]

\begin{abstract}
\vspace{-10pt}
Animating portraits using speech has received growing attention in recent years, with various creative and practical use cases. An ideal generated video should have good lip sync with the audio, natural facial expressions and head motions, and high frame quality. In this work, we present \ours, which uses speech and a single image to generate high-resolution, and expressive videos with realistic head pose, without requiring a driving video. It uses a multi-stage approach, combining the controllability of facial landmarks with the high-quality synthesis power of a pretrained face generator. \ours also allows for the control of emotions and their intensities. Our method outperforms prior methods in objective metrics for image quality and facial motions and is strongly preferred by users in pair-wise comparisons. The project website is available at \href{https://deepimagination.cc/SPACE/}{\color{Green}{\textbf{\texttt{https://deepimagination.cc/SPACE/}}}}
\end{abstract}

\section{Introduction}
\label{sec:intro}

Speech-driven portrait animation, which concerns animating a still image of a face using an arbitrary input speech signal, has a wide range of applications.
For example, it can be used for driving characters in computer games, dubbing in movies, and animating avatars for virtual assistance, virtual reality, and telecommunications.
It has to use the provided speech to predict all the nuances in human facial expressions while also guaranteeing that the animation looks natural, matches what is being said in the speech sample, and preserves the per-frame and video quality.

\begin{table*}[ht!]
\caption{\textbf{Comparison of functionality with prior works.} For fairness, we only compare with one-shot person-agnostic animation models.}
\label{table:related}
\setlength{\tabcolsep}{4pt}
\centering
\adjustbox{max width=\textwidth}{%
\begin{tabular}{L{3.1cm}ccccccccc}
\toprule
& \textbf{\ours} & Wav2Lip & MakeItTalk & PC-AVS & Audio2Head & MEAD & EAMM & GC-AVT \\
& \textbf{(ours)} & \cite{prajwal2020lip} & \cite{zhou2020makeittalk} & \cite{zhou2021pose} & \cite{wang2021audio2head} & \cite{wang2020mead} & \cite{ji2022eamm} & \cite{liang2022expressive} \\ \midrule
Controllability & \\
\hspace{4pt} $\bullet$ Emotion & \color{Green}\cmark & \xmark & \xmark & \xmark & \xmark & \cmark & \cmark & \cmark \\
\hspace{4pt} $\bullet$ Facial landmarks & \color{Green}\cmark & \xmark & \cmark & \xmark & \xmark & \xmark & \xmark & \xmark \\
\hspace{4pt} $\bullet$ Pose transfer & \color{Green}\cmark & \cmark & \xmark & \cmark & \xmark & \xmark & \cmark & \cmark \\
\hspace{4pt} $\bullet$ Pose generation & \color{Green}\cmark & \xmark & \cmark & \xmark & \cmark & \xmark & \xmark & \xmark \\
Output resolution & \textbf{\color{Green}512} & 96 & 256 & 224 & 256 & 384 & 256 & 256 \\

\bottomrule
\end{tabular}}
\end{table*}

These requirements make the task of speech-driven portrait animation challenging. To make things harder, there exist a large number of languages and facial structures that can be provided as input, each with its own unique characteristics. Furthermore, the mapping from a given input speech to corresponding facial motions is inherently one-to-many due to variations in head poses, emotions, and expressions. Despite these challenges, the ability to control each of these aspects is vital. For example, a video game character animation application might prefer the generation of exaggerated expressions and head poses, while a newscaster animation application would prefer a neutral expression. In addition, natural motions and high-resolution outputs are desirable to provide the best end-user experience. Unfortunately, no prior framework supports the ability to control emotions, facial landmarks, and poses in a single framework while producing a high-resolution output video, as summarized in Table~\ref{table:related}.

In this work, we present \ours ---a method for {\bf S}peech-driven {\bf P}ortrait {\bf A}nimation with {\bf C}ontrollable {\bf E}xpression. \ours decomposes the task into several subtasks that allow for better interpretability and fine-grained controllability. Given an input speech and a facial image, we first predict facial landmark motions in a normalized space. Operating in the facial landmark space gives us the ability to modify facial features and add actions such as blinking when desired.
Next, we apply the desired head pose to the facial landmarks and transform them into a latent keypoint space used by our pretrained face image generator~\cite{wang2021one}. This unsupervisedly-learned latent keypoint space has been shown to produce better synthesis quality than using conventional facial landmarks~\cite{siarohin2019first, wang2021one, siarohin2021motion}.
Finally, we feed these per-frame latent keypoints to our generator and produce an output video at 512$\times$512 resolution.
\ours also introduces emotion conditioning, enabling control over the emotion types and intensities in the generated video.

Even though previous approaches have also followed similar strategies wherein the audio is mapped to an intermediate representation such as facial landmarks~\cite{zhou2020makeittalk} or latent keypoints~\cite{wang2021audio2head, ji2022eamm}, no previous work has utilized both facial landmarks and latent keypoints simultaneously as intermediate face representations. By using \emph{both} explicit and latent keypoints, we are able to leverage the interpretability and direct controllability of explicit landmarks while also taking advantage of better motion transfer and image quality obtained with latent keypoints and a pretrained generator.
The main contributions of our work are as follows:
\begin{itemize}
    \item We achieve state-of-the-art quality for speech-driven portrait image animation. \ours provides better quality in terms of FID and landmark distances compared to previous methods while also generating higher-resolution output videos.
    \item Our method can produce realistic head poses while also being able to transfer poses from another video. It also offers increased controllability by utilizing facial landmarks as an intermediate stage, allowing manipulations such as blinking, eye gaze control, \etc.
    \item For the same set of inputs, our method allows for the manipulation of the emotion labels and their corresponding intensities in the output video.
\end{itemize}

\section{Related work}
\label{sec:related_work}

Animating faces using speech signals was first explored by M.\ Brand in the 1990s, who presented a hidden Markov model-based approach for \textit{Voice Puppetry}~\cite{brand1999voice}. More recently, researchers have proposed multiple methods for the task of speech-driven facial animation using deep neural networks (DNNs). Below, we discuss prior works in audio-based talking-head synthesis, video-driven facial animation, and emotion manipulation in images. A high-level comparison of current methods is provided in Table~\ref{table:related}.

\smallskip
\noindent\textbf{Speech-driven facial animation.} Approaches in this area learn to map the input speech to facial representations. Some methods animate 3D models of faces such as meshes or standard FACS-based face rigs~\cite{taylor2017deep,karras2017audio,fan2022faceformer}, whereas others directly animate raw images of faces~\cite{wang2021audio2head,zhou2020makeittalk,duarte2019wav2pix}.

Among methods that animate 3D models of faces, Taylor~\etal~\cite{taylor2017deep} first obtain phoneme sequences from audio and learn a mapping from phonemes to the mouth pose of a face model, which can be easily edited and re-targeted to other faces by animators. Zhou~\etal~\cite{zhou2018visemenet} propose VisemeNet, which predicts JALI-based visemes in a multi-stage approach. Most recently, the transformer architecture has also been leveraged for the task by Fan~\etal~\cite{fan2022faceformer}. Karras~\etal~\cite{karras2017audio} animate 3D vertices of a face given a speech signal by utilizing an end-to-end deep network consisting of a formant analysis network, an articulation network, and a learned emotion embedding. Unfortunately, these methods require special training data, such as 3D face models, which may not be available for many applications.

Another set of approaches learns to directly animate 2D face images.
Zhou~\etal~\cite{zhou2019talking} disentangle identity information and speech content and use adversarial training for learning a disentangled joint audio-visual representation.
Finally, speech and identity embeddings are used to condition a temporal GAN-based generator~\cite{vougioukas2018temporalgans}. 
Prajwal~\etal~\cite{prajwal2020lip} propose Wav2Lip, which focuses on re-dubbing videos by generating realistic lip motions by using a strong lip-sync discriminator to penalize incorrect lip shapes. However, their outputs lack realism when used with a single input image as the remainder of the face remains stationary. 
Zhou~\etal~\cite{zhou2020makeittalk} present MakeItTalk, which leverages a voice conversion model to extract disentangled speech content and speaker identity features and learns to animate facial landmarks of a given portrait or cartoon image in a speaker-aware fashion. 
This line of research, while being able to generate reasonable lip motions, lacks the ability to control head poses since no 3D is involved.
To circumvent this problem, Zhou~\etal~\cite{zhou2021pose} propose Talking-Face PC-AVS, a method that controls the head pose of a talking face by disentangling identity, pose, and speech content, with the drawback that a driving video is required for pose control.

Our method combines the benefits of both approaches by using a two-stage prediction framework. We first predict pose-normalized facial landmarks for each time step given the input audio. This gives us control over the individual landmarks, as required for fine-grained editing. Further, any rotation and translation can be applied to these landmarks with either predicted or input poses. This allows us to have full control over the 3D head poses, as well as individual landmarks, while using only a single 2D image instead of sophisticated 3D face models.

More recently, the focus has expanded to control the emotion of the output subject. Wang~\etal~\cite{wang2020mead} collect the MEAD dataset and present a method to condition talking head generation on emotion labels.
Ji~\etal~\cite{ji2021audio} extend the work to disentangle audio and emotion and apply the speech content and emotion to an input video. These methods are identity-specific and do not work in the one-shot setting.
Ji~\etal~\cite{ji2022eamm} present a one-shot generation method that disentangles emotions by utilizing a driving emotion video, which is used to guide expressions of the source image being animated.
Our method is able to extend emotional control to the one-shot setting without the need for additional emotion video input.

\smallskip
\noindent\textbf{Video-driven facial animation.} Facial animation using videos has seen profound success recently. Early works primarily focused on single subjects, where each network can only handle the specific subject seen during training~\cite{thies2015real,thies2016face2face,suwajanakorn2017synthesizing,thies2019deferred,gafni2021dynamic}. These models usually generate high-quality results, but have limited use cases since they do not easily extend to new subjects.
More recent works can perform subject-agnostic facial animation based on motions from another video~\cite{pumarola2018ganimation,geng2018warp, zhou2019talking, chen2019hierarchical, song2019talking, jamaludin2019you, vougioukas2019realistic, wang2019few, burkov2020neural, ha2020marionette, mallya2022implicit}.
For example, the first-order motion model~\cite{siarohin2019first} learns 2D latent keypoints and uses them to predict affine motions to animate the image. Face-vid2vid~\cite{wang2021one} learns 3D latent keypoints to warp the image, with control over the output poses. However, these methods rely on another video to borrow driving motions from, while we only need audio input to animate an image.

In our framework, we map the input audio to the latent space of face-vid2vid~\cite{wang2021one}, the \sota video-driven facial animation network, which is fixed and focuses on synthesizing the final images. Using a pretrained face synthesis network gives us several benefits. First, the training time is largely reduced as we only need to focus on generating latent keypoints. Second, it allows us to synthesize a high-resolution output at 512$\times$512 pixel resolution. Finally, by modulating the inputs with emotion labels, we learn the ability to control emotion and emotion intensity.

\begin{figure*}[ht!]
\centering
    \includegraphics[width=\textwidth,trim={0 1.5cm 2.25cm 0},clip]{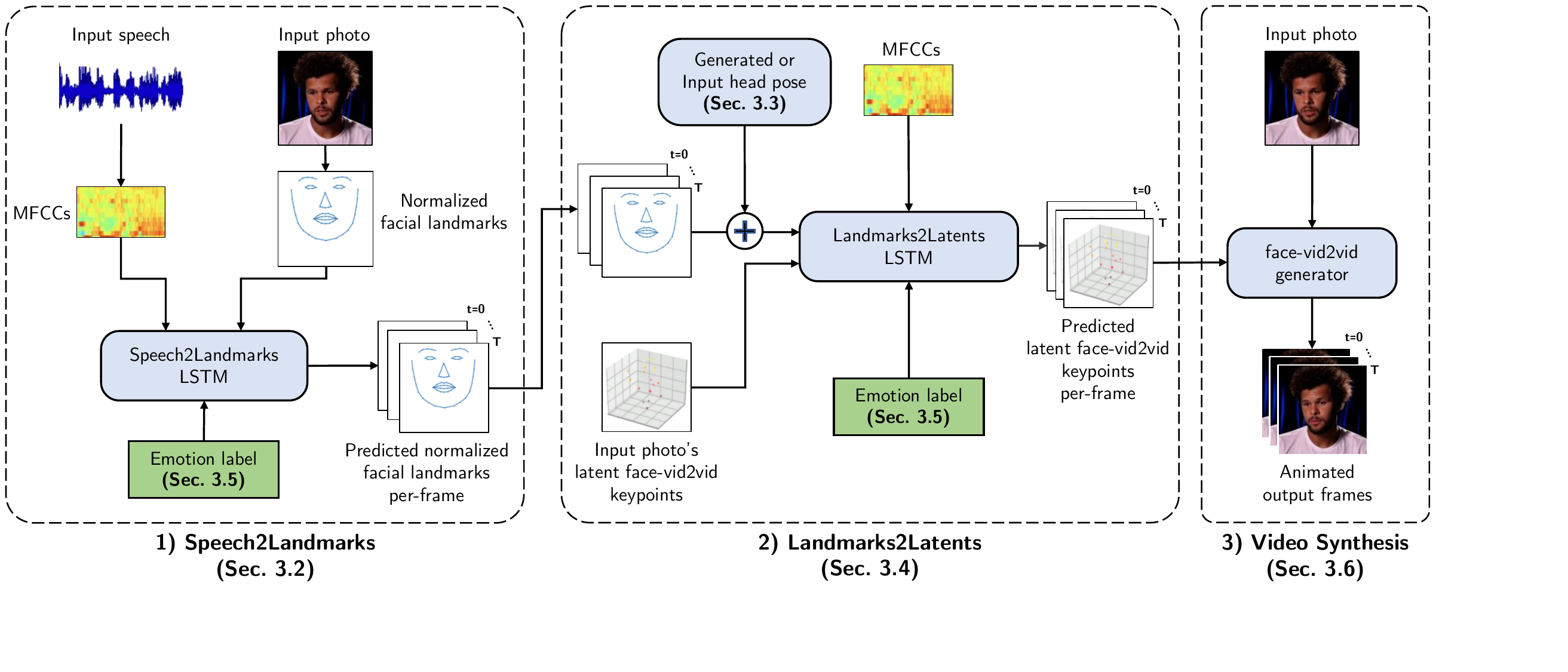}
    \caption{{\bf Overview of \ours.} Our framework consists of three stages. The first stage, Speech2Landmarks, predicts per-frame normalized facial landmarks given the initial facial landmarks and input speech. The second stage, Landmarks2Latents, predicts the latent face-vid2vid keypoints for each frame, using the input audio and previously predicted facial landmarks. The last stage warps the input image using the latent keypoints to produce animated output frames at 512$\times$512 px. Note that the intermediate predictions allow for increased controllability -- facial landmarks can be modified to add eye blinking and change head pose, while latent keypoints can be used to modulate emotions.}
    \label{fig:overview}
\end{figure*}

\section{Method}
\label{sec:method}

\ours takes an input speech clip and a face image, as well as (optionally) an emotion label, and produces an output video. It decomposes this task into three stages: 
\setlist{nolistsep}
\begin{enumerate}[left=0pt,noitemsep]
    \item Speech2Landmarks (Sec.~\ref{subsec:speech2landmarks}): Given an input image, it extracts normalized 3DDFA~\cite{zhu2017face,guo2020towards} and MTCNN~\cite{zhang2016joint} facial landmarks, and predicts their per-frame motions based on the input speech and emotion label.
    \item Landmarks2Latents (Sec.~\ref{subsec:landmarks2latents}): This step translates the per-frame posed facial landmarks into latent keypoints used by face-vid2vid~\cite{wang2021one}, a pretrained image-based facial animation model.
    \item Video synthesis (Sec.~\ref{subsec:video_synthesis}): Given the input image and the per-frame latent keypoints predicted in the previous step, the face-vid2vid generator outputs an animated video.
\end{enumerate}

\smallskip\noindent
This decomposition has multiple advantages. First, it allows for fine-grained control on the output facial expressions. For example, facial landmarks can be modified to introduce eye blinking or manipulated to apply the desired head pose, either using provided pose inputs or predicted poses (Sec.~\ref{subsec:posegen}). Further, latent keypoints can be modulated with emotion labels (Sec.~\ref{subsec:excon}) to change the expression intensity, as well as control the gaze direction. By leveraging a pretrained face generator, we are able to reduce the training cost, as well as obtain high-quality output videos. The overall framework is illustrated in Fig.~\ref{fig:overview}, and components are described below.

\subsection{Preliminaries}
\label{sec:method:background}

\noindent\textbf{Pretrained talking-head animation model.} Instead of learning an image generator from scratch, we rely on pretrained face-vid2vid~\cite{wang2021one}: a \sota framework for animating a single source image using motions from a chosen driving video. To transfer motions from a driving video to a source image, it relies on unsupervisedly-learned latent keypoints extracted from the input image and video frames.

For each input frame, the face-vid2vid encoder predicts $20$ latent keypoints. Given the latent keypoints of the input source image ($\mathrm{kp}_s$) and the current driving video frame ($\mathrm{kp}_t$), the decoder predicts a flow-based warping field. By applying the warping field to the source image features, it produces an output image that contains the identity of the source image in the pose of the driving frame. Further details are available in the supplementary and the face-vid2vid paper~\cite{wang2021one}.

As described later in Sec.~\ref{subsec:landmarks2latents}, we predict latent keypoints per frame given the input (source) image and speech. Using source image features and keypoints, along with the predicted per-frame keypoints, the face-vid2vid decoder produces an output image for each time step. This allows us to produce outputs at 512$\times$512 resolution, higher than all prior works on speech-based photo animation.

\begin{figure}[ht!]
\centering
    \includegraphics[width=\columnwidth,trim={0 6.2cm 7.3cm 0},clip]{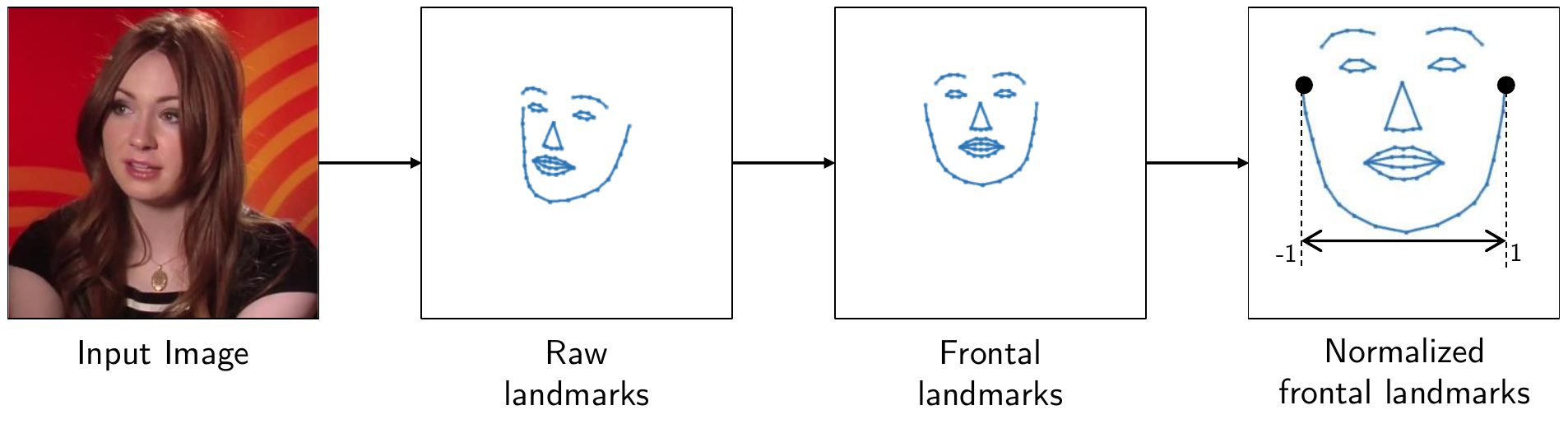}
    \caption{{\bf Input data preprocessing.} In order to obtain normalized facial landmarks per-frame for training, we run 3DDFA on each image to obtain 3D raw landmarks and the head pose (rotation and translation). We obtain frontal landmarks by undoing the rotation. Finally, we normalize the landmarks such that the distance between the landmarks of the left ear and the right ear is 2. }
    \label{fig:landmarks_processing}
\end{figure}
\begin{figure*}[ht!]
\adjustbox{max width=\textwidth}{
\centering
\setlength{\tabcolsep}{1pt}
\begin{tabular}{C{0.2\textwidth}C{0.2\textwidth}C{0.2\textwidth}C{0.2\textwidth}C{0.2\textwidth}}
    \includegraphics[width=\linewidth, height=\linewidth]{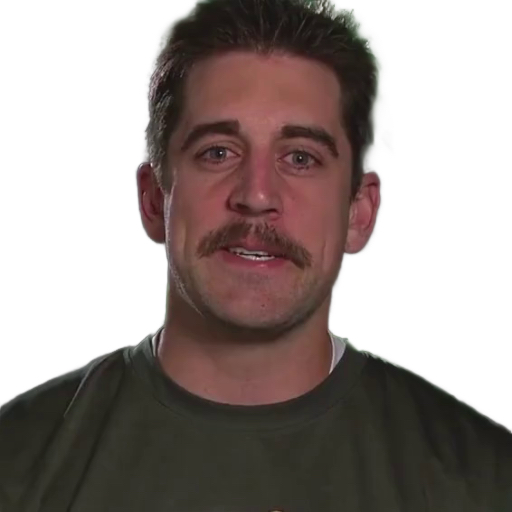} & \multicolumn{4}{c}{\href{https://deepimagination.cc/SPACE/data/intermediates/id00017-M6PYYNz3pac-00033.mp4}{\includegraphics[width=0.8\textwidth]{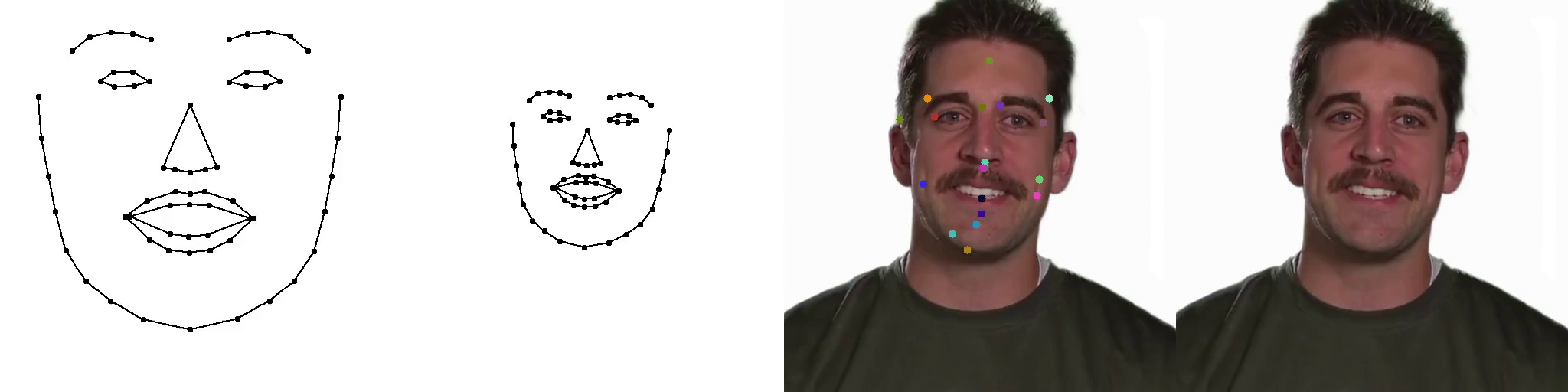}}} \\
    \includegraphics[width=\linewidth, height=\linewidth]{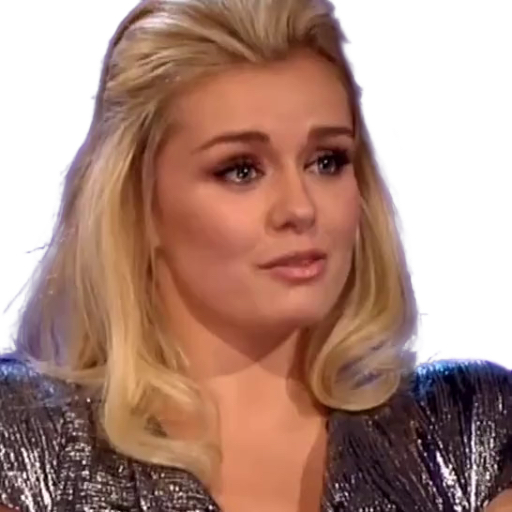} & \multicolumn{4}{c}{\href{https://deepimagination.cc/SPACE/data/intermediates/id01460-2vJlUcWI2DQ-00012.mp4}{\includegraphics[width=0.8\textwidth]{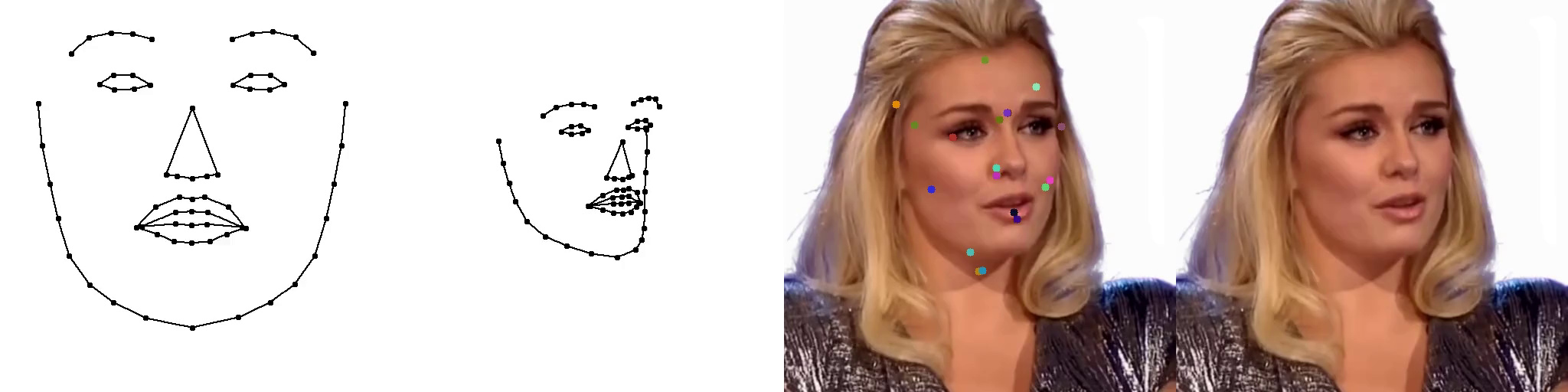}}} \\
    (a) Input image & (b) Predicted normalized facial landmarks & (c) Predicted posed facial landmarks & (d) Predicted latent face-vid2vid keypoints & (e) Animated output frames
\end{tabular}}
\caption{{\bf Inputs, predicted intermediate and final outputs, for a fixed pose.} In the first Speech2Landmarks (Sec.~\ref{subsec:speech2landmarks}) stage, given an input image (a) and speech, our method predicts facial landmark motions in the normalized space (b). These predictions are scaled, rotated, and translated to the required pose by PoseGen (Sec.~\ref{subsec:posegen}). In these samples, we transform them back to the input image space (c). In the Landmarks2Latents stage (Sec.~\ref{subsec:landmarks2latents}), these facial landmarks are used to predict corresponding face-vid2vid latent keypoints (d) (which are shown overlayed on the final output image). In the final Video Synthesis stage (Sec.~\ref{subsec:video_synthesis}), the face-vid2vid generator produces the final outputs (e) using the latent keypoints and the input image. \acrobat}
\label{fig:sample_outputs}
\end{figure*}

\smallskip
\noindent{\bf Dataset preprocessing.} Given a talking-head video, we first extract per-frame facial landmarks. We use the 3DDFA~\cite{guo2020towards,3ddfa_cleardusk,zhu2017face} landmark detector to extract 68 3D facial landmarks and head poses from each frame. We then frontalize the 3D facial landmarks by using the estimated head pose. Specifically, we rotate the face such that the nose tip is facing straight to the camera, aligned with the camera axis. The frontalized 3D facial landmarks are then orthographically projected to 2D, and each frame is normalized such that the distance between the two ear landmarks is fixed. To obtain accurate eye landmarks, we use MTCNN~\cite{zhang2016joint}. The 52 eye landmarks are required for controlling blinks and gazes. An overview of our facial landmark normalization is shown in Fig.~\ref{fig:landmarks_processing}. Additionally, we perform various steps of data filtering to remove noisy data from our training set.

In addition to landmarks, we also extract latent keypoints per frame using the pretrained face-vid2vid encoder. This gives per-frame (facial landmarks, latent keypoints) pairs.

From the audio, we extract 40 Mel-Frequency Cepstral Coefficients (MFCCs) using a 1024 sample FFT window size at 30 fps in order to align the audio features with the video frames. We also use audio data augmentation methods such as pitch-shifting, equalization, loudness variation, \etc.

For emotion labels, we use those provided with the training dataset. If unavailable, we utilize a pretrained emotion classifier~\cite{siqueira2020efficient} to predict the emotion for each frame.

\subsection{Speech2Landmarks}
\label{subsec:speech2landmarks}

The Speech2Landmarks (S2L) network is the first step of our framework, as shown in Fig.~\ref{fig:overview}. It uses a Long Short-Term Memory (LSTM) regressor for predicting the normalized facial landmarks given the input audio MFCCs and the normalized facial landmarks of the input image. This is represented by
\begin{align}
    \mathrm{face}_t &= \text{LSTM}_{\text{S2L}}\left(\mathrm{face}_{t-1},\,\mathrm{a}_t\right), \label{eq:s2l}
\end{align}
where $\mathrm{face}_t$ and $\mathrm{a}_t$ are the facial landmarks and audio features at time $t$, respectively. We use a convolutional neural network (CNN) to encode the audio and a multi-layer perceptron (MLP) to encode the facial landmarks. We train this network using an $\mathcal{L}_1$ loss with the ground truth normalized facial landmarks, with a higher loss scale on the y-axis to more heavily penalize vertical motion errors. 
We additionally use a velocity loss, an $\mathcal{L}_1$ loss between the first-order temporal difference of ground truth and predicted landmarks.
Given source images shown in Fig.~\ref{fig:sample_outputs}~(a) and the associated speech input, outputs from our S2L model are visualized in Fig.~\ref{fig:sample_outputs}~(b).

Using facial landmarks as an intermediate representation is advantageous as it allows for the explicit manipulation of facial features. For example, we can add eye blinks by manipulating the eye landmarks. We found making predictions in the normalized space to be very important to simplify the mapping between phonemes and lip motions.

\subsection{Pose generation}
\label{subsec:posegen}

For a given input speech, many different head pose sequences are valid. To model this variation, we use the conditional variational autoencoder architecture of Greenwood~\etal~\cite{greenwood2017predicting}. 
We train an autoregressive decoder that predicts the rotation $R$ and translation $T$ for the facial landmarks. $R$ is represented by three angles: yaw, pitch, and roll. $T$ is the displacement of the 3D landmarks in the x-, y-, and z-axes. Our network, PoseGen, at time step $t$ is expressed as
\begin{align}
    R_t, T_t = \text{LSTM}_\text{pose}(a_t, z), \label{eq:posegen}
\end{align}
where $z$ is a noise sampled from $\mathcal{N}(0,1)$ during inference, and  $\mathcal{N}(\mu,\sigma)$ during training. $\mu,\sigma$ are predicted by a bidirectional LSTM encoder that jointly encodes audio and pose.

The poses, whether predicted or extracted from a reference video, are applied to the frontal normalized landmarks predicted by our S2L module. This transforms the normalized landmarks back to the image space after an appropriate scaling factor is applied. Examples of the predicted facial landmarks (transformed to a fixed pose for better visualization) are shown in Fig.~\ref{fig:sample_outputs}~(c). Time-varying pose outputs using our generated poses are shown in Fig.~\ref{fig:teaser} and the supplementary material.

\subsection{Landmarks2Latents}
\label{subsec:landmarks2latents}

The second step, Landmarks2Latents (L2L), takes in posed facial landmarks, and generates corresponding latent face-vid2vid keypoints, as shown in Fig.~\ref{fig:overview}. 
An LSTM regressor predicts latent keypoints given encoded audio features, predicted posed landmarks, latent keypoints of the input image, and the previous prediction. L2L can be expressed as 
\begin{align}
    \mathrm{kp}_t &= \text{LSTM}_\text{L2L}(R_t\cdot\mathrm{face}_{t}+T_t,\,\mathrm{a}_t,\,\mathrm{kp}_{\text{s}},\,\mathrm{kp}_{t-1}), \label{eq:l2l}
\end{align}
where $\mathrm{face}_t$ is predicted by S2L (Eq.~\ref{eq:s2l}), $R_t$ and $T_t$ are predicted by PoseGen (Eq.~\ref{eq:posegen}), and $\mathrm{kp}_{\text{s}}$ is predicted from the input image by the face-vid2vid encoder.
For training, we use the (facial landmarks, latent keypoints) pairs generated as described in Sec.~\ref{sec:method:background}. We train this network using an $\mathcal{L}_1$ loss with the ground truth latent keypoints. Examples of the predicted latent keypoints are shown in Fig.~\ref{fig:sample_outputs}~(d), overlayed on the final outputs.

\begin{figure}[t!]
\centering

\begin{minipage}{.32\columnwidth}
    \centering
    \begin{tabular}{C{\textwidth}}
        \includegraphics[width=\textwidth, height=\textwidth]{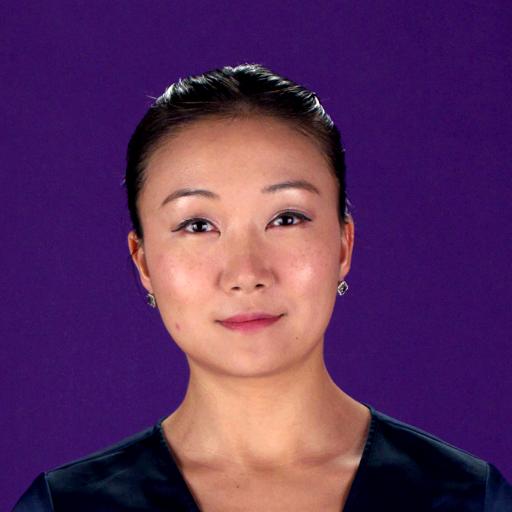} \\
        (a) Input image
    \end{tabular}
\end{minipage}
\hfill
\begin{minipage}{0.65\columnwidth}
    \centering
    \setlength{\tabcolsep}{2pt}
    \adjustbox{max width=\textwidth}{
    \begin{tabular}{C{0.5\textwidth}C{0.5\textwidth}}
        (b) 0.5 Happy & (c) 1.0 Happy \\
        \href{https://deepimagination.cc/SPACE/data/emotion/video.mp4}{\includegraphics[width=0.5\textwidth]{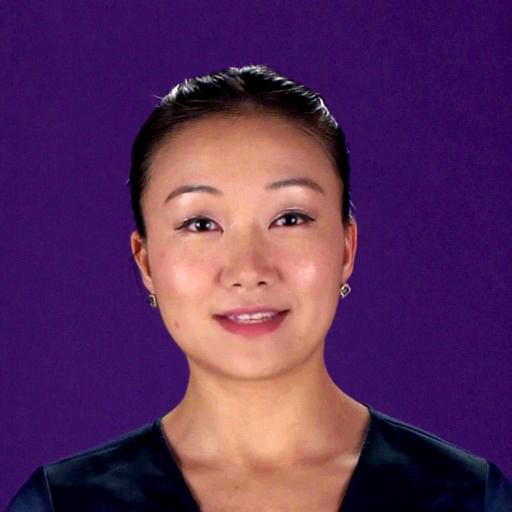}} &
        \href{https://deepimagination.cc/SPACE/data/emotion/video.mp4}{\includegraphics[width=0.5\textwidth]{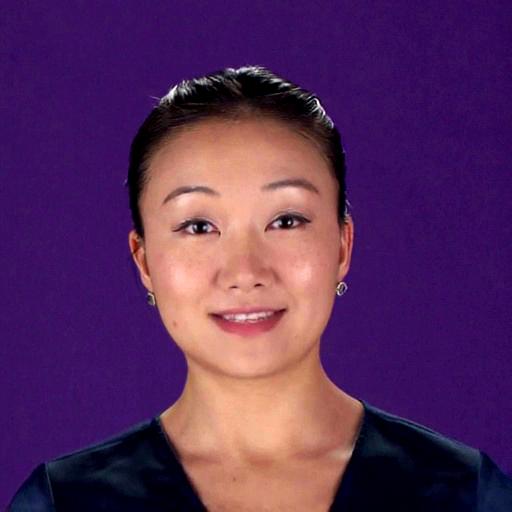}} \\
        \href{https://deepimagination.cc/SPACE/data/emotion/video.mp4}{\includegraphics[width=0.5\textwidth]{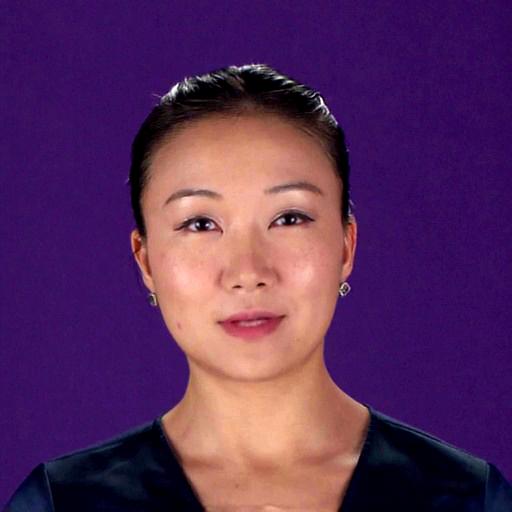}} &
        \href{https://deepimagination.cc/SPACE/data/emotion/video.mp4}{\includegraphics[width=0.5\textwidth]{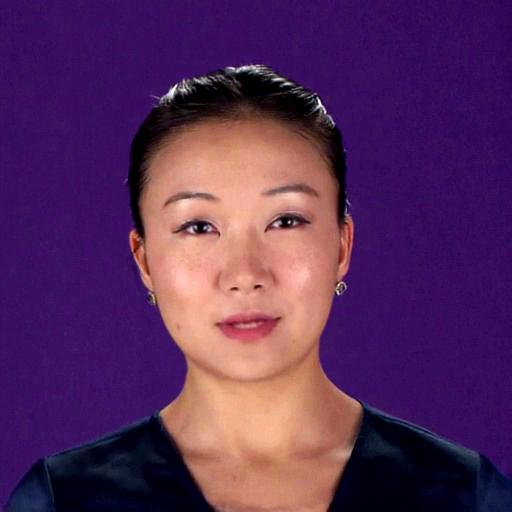}} \\
        (d) 0.5 Angry & (e) 1.0 Angry
    \end{tabular}}
\end{minipage}

\captionof{figure}{{\bf Emotion label and intensity control.} We can condition outputs on both the emotion label and its intensity. \acrobat}
\label{fig:emotion_control}

\end{figure}
\subsection{Emotion control}
\label{subsec:excon}

To provide additional user control on the generated video, we condition both S2L and L2L on the emotion of the video frames. This is achieved using Feature-wise Linear Modulation (FiLM) layers~\cite{perez2018film}. For the S2L network, we use FiLM to modulate the audio features and the initial landmark input. For L2L, we apply FiLM on the audio, landmarks, and the initial latent keypoint input. As mentioned earlier in Sec.~\ref{sec:method:background}, we use one-hot emotion labels when provided by the dataset, and a predicted per-frame probability distribution over the emotions~\cite{siqueira2020efficient} when the ground-truth emotions are unavailable. At inference, we can provide the desired combination of emotion labels and their intensities as input.

Even though we do not impose any constraints to disentangle the emotion label from the input speech, unlike prior work~\cite{wang2020mead,ji2021audio,ji2022eamm}, we find that our model is able to change facial emotions, as shown in Fig.~\ref{fig:emotion_control}. We believe there are two main reasons this simple modulation works -- 1) Our audio encoder is shallow and unidirectional. Due to a small lookahead, various augmentations, and lack of bidirectionality, it is unable to extract emotions from the audio; 2) Thus, relying on emotion-conditioned modulations of intermediate landmarks and latent keypoints, which directly control facial expressions, becomes the most effective way to produce the required expressions.

\subsection{Video synthesis}
\label{sec:method:video}
\label{subsec:video_synthesis}
The third and final step, as shown in Fig.~\ref{fig:overview}, is the talking-head video generation by the pretrained face-vid2vid~\cite{wang2021one} generator. The generator uses the input image and the per-frame latent keypoints $\mathrm{kp}_t$ predicted by the previous step, L2L (Eq.~\ref{eq:l2l}), and outputs video frames at 512$\times$512 pixel resolution. Examples of the final outputs and intermediate predictions are shown in Fig.~\ref{fig:sample_outputs}.\\

To summarize, given an input image and audio, our method learns to produce valid sequences of face-vid2vid latent keypoints. These latent keypoints are used to animate the input image using the pretrained face-vid2vid generator. In order to provide additional user control, such as blinking and pose change, we use facial landmarks as an interpretable intermediate prediction. Lastly, the emotion conditioning helps change the emotion and intensity of output expressions. Additional architectural and implementation details are available in the supplementary material.

\section{Results}

In this section, we discuss the datasets, metrics, and various experiments we conduct to validate our method.

\medskip
\noindent\textbf{Datasets.}
We train our models using three different datasets: VoxCeleb2~\cite{chung2018voxceleb2}, RAVDESS~\cite{livingstone2018ryerson}, and MEAD~\cite{wang2020mead}. {VoxCeleb2} is made up of a large collection of YouTube videos of celebrities speaking in various settings, such as interviews, speeches, and podcasts. It consists of a diverse set of speakers with a wide range of languages and accents. RAVDESS is a smaller-scale dataset compared to VoxCeleb2, consisting of approximately 7k clean video recordings of various actors speaking two sentences, with $8$ different emotions in two levels of emotional intensity. MEAD is a larger-scale emotional dataset consisting of 40 hours of audio-visual recordings from 60 actors (48 of which are made publicly available). It consists of multi-camera recordings of actors speaking thirty sentences with $8$ emotions in three levels of intensity.

After combining, preprocessing, and filtering the three datasets as described in Sec.~\ref{sec:method:background}, we obtain $\sim$270k training videos. For evaluation, we manually select and set aside 350 videos of unseen identities. This subset is selected to ensure it captures a diverse set of speakers with different initial poses. More details of the datasets are in the supplementary.

\medskip
\noindent\textbf{Baselines.} We compare \ours to 3 prior works on speech-driven animation --
1) \textbf{Wav2Lip}~\cite{prajwal2020lip},
2) \textbf{MakeItTalk}~\cite{zhou2020makeittalk}, and
3) \textbf{Talking Face-PC-AVS (PC-AVS)}~\cite{zhou2021pose}.
We chose these baselines for two main reasons. First, the source code and pretrained model weights are available. Second, they represent the state-of-the-art talking-head video generation using arbitrary identities. These methods, like ours, do not need to be fine-tuned or trained separately for a new identity that was not seen during training.  The publicly available code for MEAD~\cite{wang2020mead}, which has emotion control, does not support arbitrary identities for inference.
Note that Wav2Lip and PC-AVS both require an input video to determine the pose of the generated output. However, we provide all models with only two inputs in our setting: an initial photo and the driving speech. Therefore, these two methods will naturally only generate videos with a single pose. Thus, we use a fixed pose with \ours for a fair comparison.

\medskip
\noindent\textbf{Metrics.} We use the following metrics to measure facial landmark motion and photorealism:
\begin{itemize}
    \item \textbf{Lip sync quality}: We use the lip sync confidence score output by SyncNet~\cite{Chung2016OutOT} as a proxy for lip sync quality.
    \item \textbf{Landmark accuracy}:
    We extract facial landmarks from outputs, frontalize, and normalize them such that the left and right edge of the mouth landmarks are at $(-1,0)$ and $(1,0)$, respectively. We measure the mean absolute error (MAE) between the predicted and ground truth mouth landmarks positions (M-P) and velocities (M-V). We compute errors in face geometry (F-P) and velocity (F-V), using the normalization from Sec.~\ref{sec:method:background}.
    \item \textbf{Photorealism}: We measure the Frechet Inception Distance (FID)~\cite{heusel2017gans} of outputs with the ground truth.
    \item \textbf{Human evaluation}: We perform a forced preference A/B test with the four baseline models for the same input image and speech audio. Each pair is rated by 3 users, and we report the average preference score.
\end{itemize}

Additional details about the metrics, including the human evaluation interface, are available in the supplementary.

\begin{table}[t!]
\caption{\textbf{Quantitative evaluation results.}
$\downarrow$ indicates lower is better. Although PC-AVS obtains a better SyncNet score, we found the score to be highly sensitive to the input crop. In fact, the mean score on real videos is $4.85$.}
\label{tab:vs_baselines}
\adjustbox{max width=\columnwidth}{%
\begin{tabular}{lcccccc}
\toprule
& FID $\downarrow$   & M-P $\downarrow$    & M-V $\downarrow$   & F-P $\downarrow$   & F-V $\downarrow$  & Sync \\
\midrule
PC-AVS~\cite{zhou2019talking} & 66.68 & 0.032 & 0.013 & 0.024 & 0.004 & \textbf{7.00}\\
MakeItTalk~\cite{zhou2020makeittalk} & 30.68 & 0.035 & 0.012 & 0.020 & 0.004 & 2.61\\
Wav2Lip~\cite{prajwal2020lip} & 15.67 & 0.030 & 0.013 & 0.017 & 0.004 & 5.08\\
\midrule
\multicolumn{7}{c}{Ablations} \\
S2Latent & 11.82 & \textbf{0.021} & \textbf{0.005} & 0.013 & 0.003 & 2.94\\
L2L-landmarks & 15.63 & 0.026 & 0.007 & 0.008 & 0.002 & 2.78\\
\midrule
\ours (ours) & \textbf{11.68} & 0.025 & 0.007 & \textbf{0.008} & \textbf{0.002} & 3.61\\
\bottomrule
\end{tabular}
}
\end{table}

\begin{table}[t!]
\caption{{\bf User preference scores.} We perform a forced preference test where users are presented with two videos with audio. They are asked to select which one of the two is more realistic, with a focus on the face and mouth region. We obtain $\sim$900 ratings per pair of models.
Users overwhelmingly prefer \ours.
}
\label{tab:user_study}
\adjustbox{max width=\columnwidth}{%
\begin{tabular}{lccc}
\toprule
& PC-AVS~\cite{zhou2019talking} & MakeItTalk~\cite{zhou2020makeittalk} & Wav2Lip~\cite{prajwal2020lip} \\
\midrule
\ours (ours) & 89.1\% & 87.8\% & 72.6\% \\ 
\bottomrule
\end{tabular}
}
\end{table}

\begin{figure*}[ht!]
    \adjustbox{max width=\textwidth,center}{
    \setlength{\tabcolsep}{0pt}
    \centering
    \hspace{-8pt}
    \begin{tabular}{C{0.2\textwidth}C{0.2\textwidth}C{0.2\textwidth}C{0.2\textwidth}C{0.2\textwidth}}

        \includegraphics[width=\linewidth, height=\linewidth]{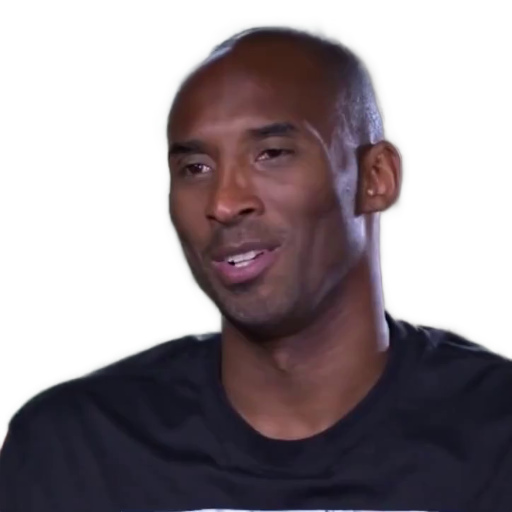} &
        \multicolumn{4}{c}{\href{https://deepimagination.cc/SPACE/data/comparison/id04862-4r-u7YAGLoA-00039.mp4}{\includegraphics[width=0.8\textwidth]{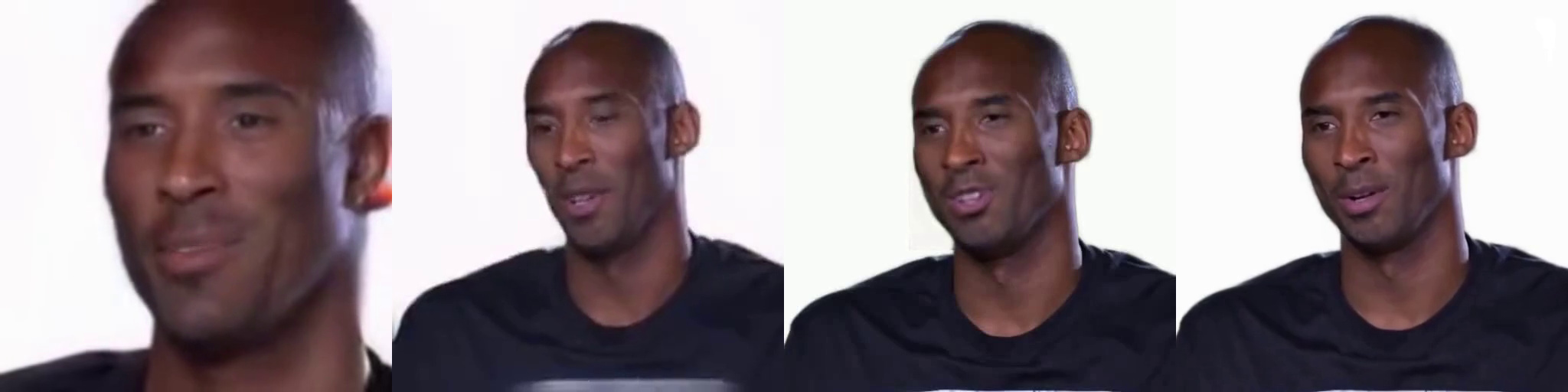}}} \\[-4pt]

        \includegraphics[width=\linewidth, height=\linewidth]{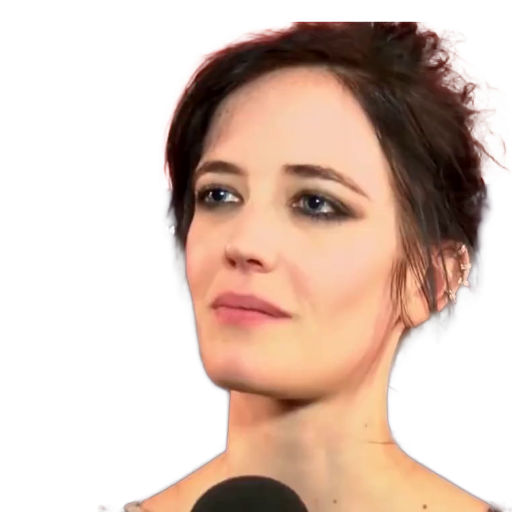} &
        \multicolumn{4}{c}{\href{https://deepimagination.cc/SPACE/data/comparison/id10306-5jsUg3uxG-E-00005.mp4}{\includegraphics[width=0.8\textwidth]{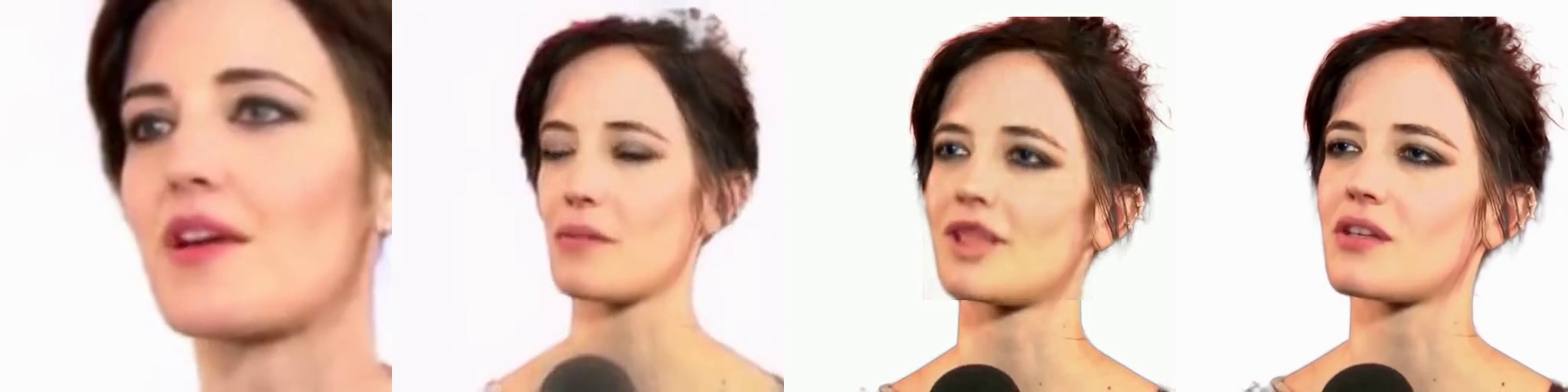}}} \\[-4pt]

        \includegraphics[width=\linewidth, height=\linewidth]{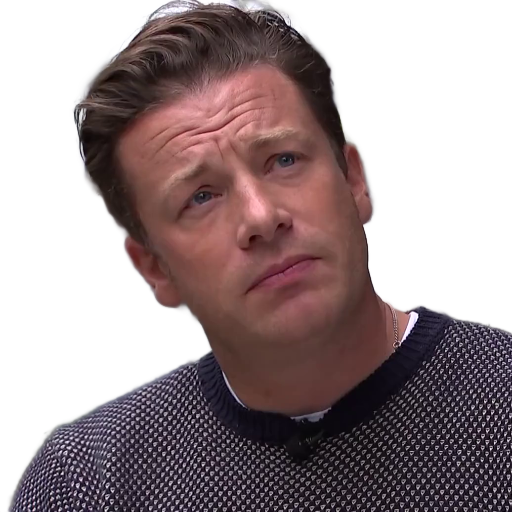} &
        \multicolumn{4}{c}{\href{https://deepimagination.cc/SPACE/data/comparison/id03789-9alL5EAVFZU-00114.mp4}{\includegraphics[width=0.8\textwidth]{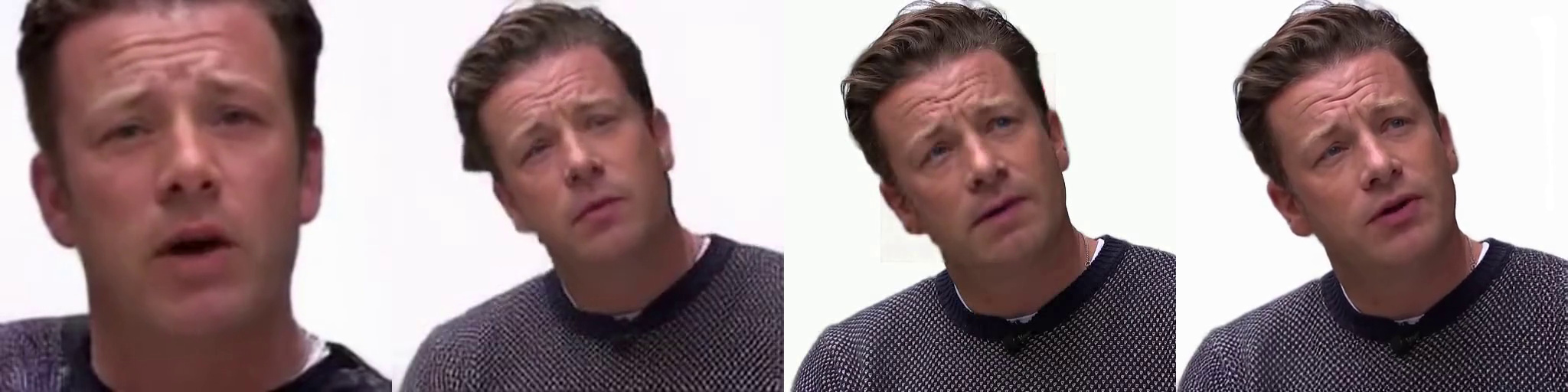}}} \\

        (a) Input image & (b) PC-AVS~\cite{zhou2019talking} & (c) MakeItTalk~\cite{zhou2020makeittalk} & (d) Wav2Lip~\cite{prajwal2020lip} & (e) \ours (ours)
    \end{tabular}}
    \caption{{\bf Comparisons with prior works.} \ours is able to animate faces in any pose, with better temporal stability and per-frame quality. Note the temporal jittering in (b), deformed face shapes in (c), and unnatural mouth motions in (d). \acrobat
    }
    \label{fig:comparison}
\end{figure*}

\medskip
\noindent\textbf{Quantitative results.}
As shown in Table~\ref{tab:vs_baselines}, \ours achieves the lowest FID, indicating that our method produces the best image quality. We also outperform the baselines on the normalized landmark distance metrics. Despite obtaining better output quality and lower temporal jittering, our method achieves poorer SyncNet scores. We found that SyncNet scores are very sensitive to the input crop and suspect that our larger-crop outputs may have resulted in worse scores. Surprisingly, the mean SyncNet score on real videos is 4.85, lower than Wav2Lip at 5.08 and PC-AVS at 7.00.

\medskip
\noindent\textbf{Qualitative results.}
Fig.~\ref{fig:comparison} shows examples of videos generated by various methods. While previous methods work on frontal-facing and closely-cropped input images, they suffer from degraded quality or fail for arbitrary poses and larger crops.
Since we predict mouth motions in the normalized landmark space and then apply pose transformations, we are able to handle poses from face images in the wild.
In addition, \ours is also able to generate missing details such as teeth, while other methods either fail or introduce artifacts. We also found that our method adds realistic head and shoulder motions when the audio has a breathing sound.
We find that \ours has smoother lip motion compared to Wav2Lip and PC-AVS, where the motions tend to be more exaggerated. We suspect that this is another reason for poorer SyncNet confidence scores for our method.

As seen in Table~\ref{tab:user_study}, users overwhelmingly prefer \ours over other methods. This study asks users to rate which of the two videos is more realistic in a two-choice test.

\medskip
\noindent\textbf{Ablations.} To isolate the contributions of various design choices, we perform ablation studies with three alternatives -- 1) S2Latent: directly predicting latent keypoints from speech instead of the proposed multi-stage system, 2) L2L-landmarks: S2L uses audio, but L2L only uses landmarks, and 3) S2L-raw: using raw landmarks to train S2L.

From Table~\ref{tab:vs_baselines}, it can be seen that S2Latent achieves strong performance on the chosen metrics; however, since it is not informed of the pose, there tends to have pose drift in the generated videos, leading to poor overall motion quality. Additionally, we are unable to control aspects such as eye blinks and mouth opening/closing in this case.
On the other hand, for L2L-landmarks we find the fine-grained motions to be missing. This can be attributed to the fact that facial landmarks do not accurately capture fine-grained mouth motions and expressions, while our full model can rely on audio for the missing details.
Finally, S2L-raw suffers from both of the issues, i.e., failing to predict accurate mouth motions, and pose drift due to missing pose information. Hence we exclude it from the quantitative evaluations.
For visual results, please see the supplementary material.

\medskip
\noindent\textbf{Novel applications.}
\ours can be used in video conferencing as sending only audio and a still image saves substantial bandwidth over sending the full video.
Furthermore, due to its pose controllability, it can be combined with existing approaches so that the user can freely switch between using video or audio inputs. In low bandwidth scenarios, the video conferencing system can fall back to the audio-driven mode and still generate realistic output videos. We show such an application with seamless video-to-audio-driven switching in the supplementary material.

\section{Discussion}
In this work, we have presented \ours, a speech-driven face animation framework that produces \sota realistic and high resolution videos.
Due to our novel intermediate representations, we also gain control over the head pose, eye blinking, and gaze directions. Our method is also able to utilize emotion labels and their intensities as input to produce diverse outputs with different emotions.
Moreover, \ours also possesses the ability to generate poses from audio or use poses from a different driving video.
The capabilities of \ours open new avenues for applications in video conferencing, gaming, and media synthesis.

\medskip
\noindent\textbf{Limitations and Societal impact.}
Our method might fail on underrepresented emotions, phonemes, and visemes. A diverse training dataset should help mitigate such issues. Extreme head poses can cause inaccuracies in 3DDFA predictions, and thus our method.
Our method has the potential for negative impact if used to create \emph{deepfakes}. Using voice cloning, a malicious actor can create videos enabling identity theft or dissemination of fake news. However, in controlled settings,
\ours can be used for positive creative purposes.

{\small
\bibliographystyle{ieee_fullname}
\bibliography{egbib}
}

\clearpage
\appendix
\appendix
\twocolumn[{%
\renewcommand\twocolumn[1][]{#1}%
\begin{center}
    \textbf{\LARGE Appendix}
    \vspace{10pt}
\end{center}%
}]

\section{Network architecture and training}

\subsection{Speech2Landmarks}

\begin{figure}[t]
    \includegraphics[width=\columnwidth]{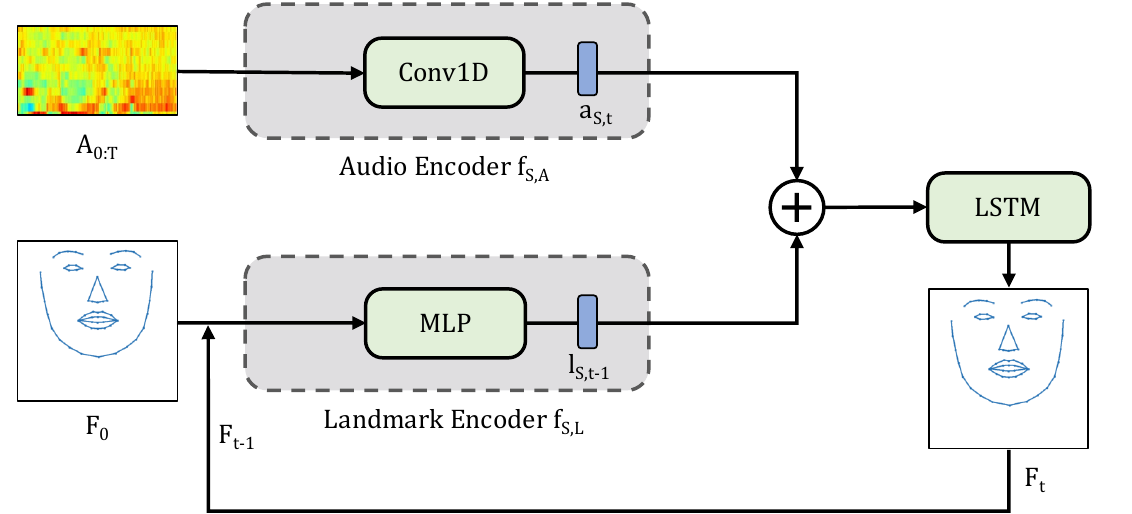}
    \caption{{\bf \speechtolandmarks architecture.} \speechtolandmarks takes as input the normalized face landmarks $F_0$, the mouth landmarks sequence $M_{1:T}$, and the MFCC features of the speech recording $A_{0:T}$. It is an autoregressive model that takes as input the predicted face landmarks from previous time step, and audio features and mouth landmarks for the current time step. It consists of a face landmarks encoder $f_{F,L}$, a mouth landmarks encoder $f_{F,E}$, an audio encoder $f_{F,A}$,
    and an LSTM network.}
    \label{fig:speech2landmarks}
\end{figure}

Fig.~\ref{fig:speech2landmarks} shows the high-level architecture of the Speech2Landmarks (S2L) model. The main components include an audio encoder, landmark encoder, and the LSTM regressor that autoregressively predicts the face landmarks frame-by-frame.

The audio encoder is a 12-layer convolutional network with 1024 1D filters. Each layer uses leaky ReLU activations and batch-norm. The first 4 layers use symmetrical filters spanning 3 time-steps. The last 8 layers use causal filters spanning 5 time-steps. The goal of this asymmetrical padding is to have the audio encoder focus more on the past and have fewer time-steps look-ahead, thus enabling use in real-time applications.

There are two landmark encoders. For the 68 3DDFA landmarks  we use an 8 layer fully-connected network with 1024 hidden units with leaky ReLU activation and batch norm. For the 52 MTCNN eye landmarks we use an 4 layer fully-connected network with 1024 hidden units with leaky ReLU activation and batch norm. Both the encoders are multi-layer fully connected networks.

The regressor is a 2-layer LSTM with 1024 hidden units. We also found that using a learned initialization for the hidden state leads to better initial outputs. For this purpose we use a 4-layer fully connected network which takes the initial time-step of all inputs:
\begin{align}
    \mathrm{h}_0 &= \text{MLP}(\mathrm{face}_{0},\,\mathrm{a}_0)\label{eq:s2l_hidden}
\end{align}

\subsection{Pose Generation}

\begin{figure}[t!]
    \includegraphics[width=\columnwidth]{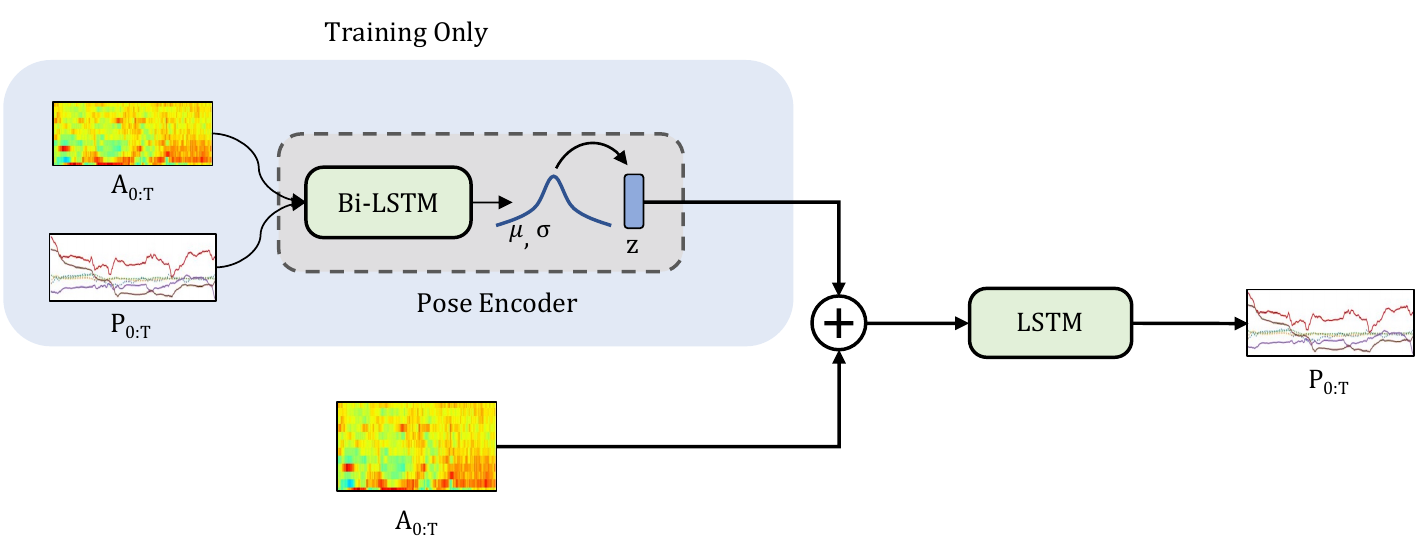}
    \caption{{\bf PoseGen architecture.} PoseGen is a variational autoencoder which jointly encoders audio and pose sequence. The decoder subsequently reconstructs the pose sequence conditioned on the input audio.}
    \label{fig:posegen}
\end{figure}

Fig.~\ref{fig:posegen} depicts the high-level architecture of our Pose Generation network (PoseGen).
The encoder is a 2-layer bi-LSTM with 256 hidden units. The latent z is 64 dimensional. The decoder is a 2-layer LSTM with 256 hidden units.

\begin{figure}[t!]
    \includegraphics[width=\columnwidth, trim={0cm 0cm 0cm 0cm}, clip]{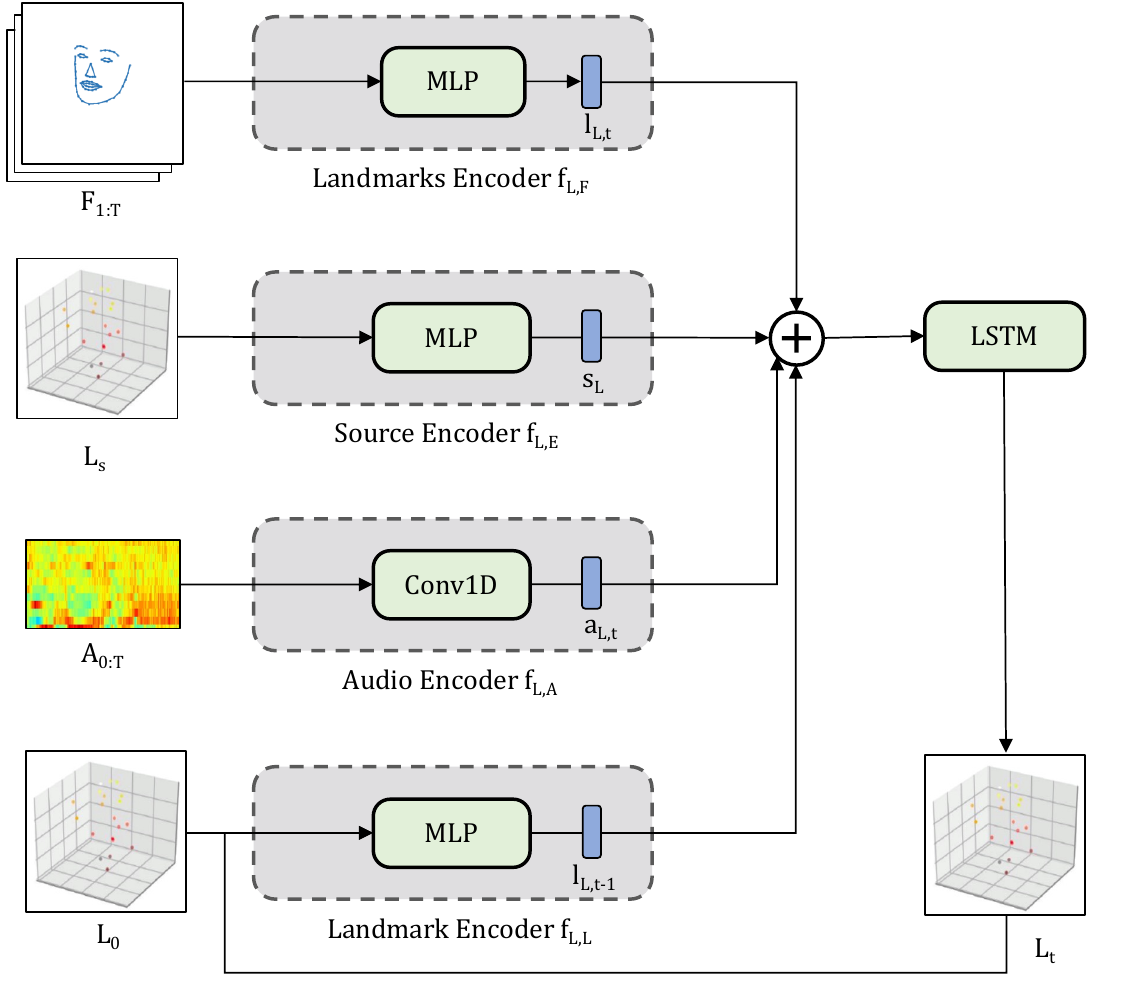}
    \caption{{\bf \landmarkstolatents architecture.} \landmarkstolatents is the final autoregressive component of \ours. It is trained to transform facial landmarks to self-supervised keypoints learned by the talking head synthesis model.  \landmarkstolatents takes as input the full face landmarks aligned with the image, which in turn are aligned with the keypoints. It also uses the input speech as an additional source of information. At a particular time step, the model takes the previous keypoints encodings, and the current landmarks and audio features.}
    \label{fig:s2kp}
\end{figure}

\subsection{Landmarks2Latents}

Fig.~\ref{fig:s2kp} shows the high-level architecture of the Landmarks2Latents (L2L) model. The main components include an audio encoder, face landmark encoder, latent keypoint encoder, and the LSTM regressor that autoregressively predicts the latent landmarks frame-by-frame. 

The audio encoder, face landmarks encoders, and the regressor follow the same architecture as Speech2Landmarks.

The source encoder encodes the latent keypoints from the source image. It is a 4-layer fully connected network with 1024 hidden units with leaky ReLU activation and batch norm. The latents encoder encodes the latent keypoints of the previous time-step. It is an 8-layer fully connected network with 1024 hidden units with leaky ReLU activation and batch norm. 

\subsection{Video Generation}

For details about face-vid2vid, please visit the project page: \url{https://nvlabs.github.io/face-vid2vid/}.

\subsection{Training details}

Each network is trained using the Adam optimizer with a batch size of 256. We use a learning schedule which warms up the learning rate from $1\mathrm{e}{-5}$ to $5\mathrm{e}{-4}$ over 10000 steps. Then we use a cosine schedule to decay the learning rate to 0 over 1 million steps. We find our networks converge within 200k steps.

\section{Data preprocessing}

Since VoxCeleb1 and VoxCeleb2 datasets were captured in the wild, there is a large variation in the quality of videos due to different recording devices, camera viewpoints, head poses, occlusions, \etc. We found it essential to filter the data to remove outliers and obtain a clean subset for training our models.
First, we remove any video where a state-of-the-art face detection algorithm fails to detect a face in any of its frames. This is because we assume facial landmarks are available for all the frames in a video. If a face detector cannot detect a face in a frame, then we will either fail to extract the facial landmarks from the frame or have unreliable facial landmarks from the frame. Neither is desirable.
We also remove a video with a large temporal difference between facial landmarks in any pair of its neighboring frames. Such a case often occurs in an edited video where the view transitions from one subject to another.
In addition, we remove videos with large head rotation and camera zoom motion.
We track the scale of the face landmarks and use the difference between the minimum and maximum scales as the criterion for filtering. 
We also remove videos with head rotation greater than $45$ degrees along any axis.
We use a pre-trained hand tracker to remove any videos where hands are detected.

For the Ryerson and MEAD datasets, we use a face detector to crop and resize the full size videos to 512$\times$512. MEAD contains videos from various camera perspectives: front, up, down, left 30 degrees, right 30 degrees, left 60 degrees and right 60 degree. We discard the left 60 degrees and right 60 degrees videos.

\subsection{Test Set details}

Our test set was chosen to have a diverse set of initial poses. Using head pose estimation, we found that our test set has $\sim$45\% data with rotations less than 15 degrees, $\sim$35\% data with rotations between 15 and 30 degrees, and $\sim$20\% data with rotations between 30 and 45 degrees.

\section{Evaluation}

\begin{figure}[t!]
    \fbox{\includegraphics[width=\columnwidth]{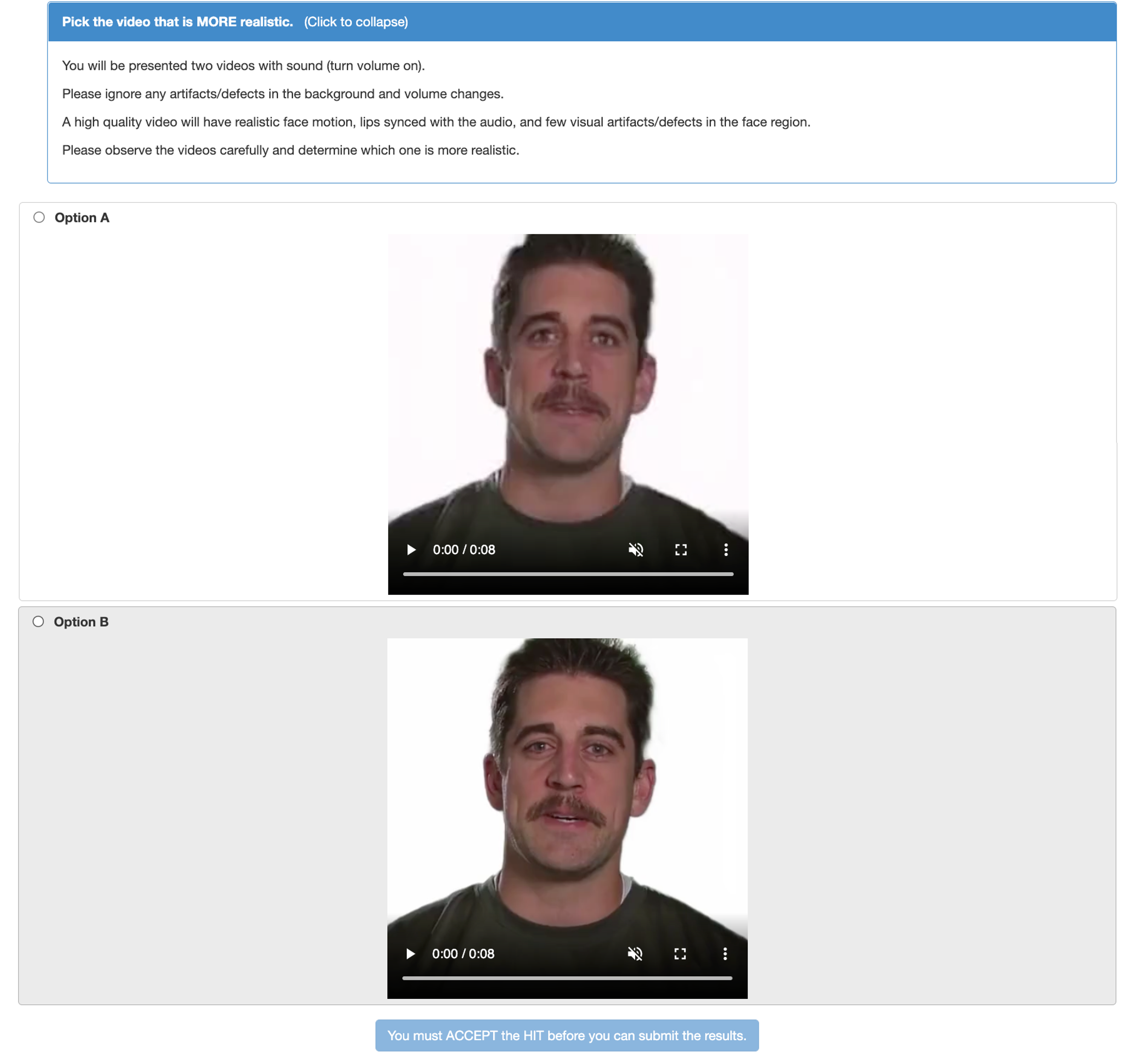}}
    \caption{{\bf Amazon Mechanical Turk interface.} We ask users to choose the more realistic video, with a focus on the face and mouth region.}
    \label{fig:amt}
\end{figure}
Fig.~\ref{fig:amt} shows the interface of the user study. As described in the text, a user is presented with a pair of videos from randomized models. Each pair consists of videos generated using same reference image and audio with different models. The user is asked to pick the video they prefer in terms of face and mouth motion.

\end{document}